\documentclass[]{style}

\usepackage[toc,page,header]{appendix}
\usepackage{enumitem}
\usepackage{hyperref}

\usepackage{natbib}

\usepackage{CJKutf8}

\usepackage{xargs}  

\usepackage{todonotes}  

\usepackage{multirow}

\usepackage{cleveref}

\usepackage{amsmath}
\usepackage{dsfont}

\usepackage{svg}

\usepackage{mathrsfs}
\usepackage{adjustbox}
\usepackage{multirow}
\usepackage{multicol}
\usepackage{tcolorbox}
\usepackage{changepage}
\usepackage{enumitem}
\usepackage{graphicx}
\usepackage{amssymb}
\usepackage{xcolor}
\usepackage{float}
\usepackage{multirow}
\usepackage{threeparttable}
\usepackage{graphicx}
\usepackage{subcaption}
\usepackage{algorithm}
\usepackage{algpseudocode}
\usepackage{wrapfig}
\usepackage[table]{xcolor}  
\usepackage{colortbl}       
\usepackage{tabularx} 
\usepackage{makecell} 
\usepackage[dvipsnames]{xcolor}

\newcolumntype{g}{>{\columncolor{gray!10}}c} 

\definecolor{catgray}{gray}{0.9}
\definecolor{skyblue}{rgb}{0.53,0.81,0.92} 

\colorlet{skyblue!30}{skyblue!30!white} 

\definecolor{customblue}{RGB}{70,130,180}  

\newtcolorbox{evolbox}[2][]{%
  enhanced,
  colframe=customblue,
  colback=white,
  coltitle=white,
  rounded corners,
  boxrule=1pt,
  titlerule=0pt,
  toptitle=1mm,
  bottomtitle=1mm,
  fonttitle=\bfseries,
  width=#2\textwidth, 
  #1
}

\usepackage{minitoc}
\usepackage{url}

\PassOptionsToPackage{table,xcdraw}{xcolor}
\usepackage{titletoc}
\usepackage{placeins}
\usepackage{pifont}

\definecolor{RowBlue}{HTML}{E9F2FB}
\definecolor{RowRed}{HTML}{F9EAEA}
\definecolor{Top1}{HTML}{50DB4B} 
\definecolor{Top2}{HTML}{A5FFA2} 
\definecolor{Top3}{HTML}{D9FFD9} 
\definecolor{Sub1}{HTML}{EAB8B8}
\definecolor{Sub2}{HTML}{E4E4E4}

\renewcommand{\emph}[1]{\textit{#1}}

\title{HumanScale: Egocentric Human Video Can Outperform Real-Robot Data for Embodied Pretraining}

\author{
  Juncheng Ma\texorpdfstring{$^{*}$}{} \hspace{0.05cm}
  Jianxin Bi\texorpdfstring{$^{*}$}{} \hspace{0.1cm}
  Yufan Deng\texorpdfstring{}{} \hspace{0.1cm}
  Xuanran Zhai\texorpdfstring{}{} \hspace{0.1cm}
  Kewei Zhang\texorpdfstring{}{} \hspace{0.1cm}
  Ye Huang\texorpdfstring{}{} \hspace{0.1cm} \\
  Bo Liang\texorpdfstring{}{} \hspace{0.1cm}
  Shukai Gong\texorpdfstring{}{} \hspace{0.1cm}
  Jiankai Tu\texorpdfstring{}{} \hspace{0.1cm}
  Xiaotian Tang\texorpdfstring{}{} \hspace{0.1cm}
  Jiaxin Li\texorpdfstring{}{} \hspace{0.1cm}
  Kaiqi Chen\texorpdfstring{}{} \hspace{0.1cm}
  Duomin Wang\texorpdfstring{}{} \hspace{0.1cm} 
  Yuqi Wang\texorpdfstring{}{} \hspace{0.1cm} 
  Bingyi Kang\texorpdfstring{}{} \hspace{0.1cm} 
  Eric Huang \hspace{0.1cm}
  Zhiyang Dou \hspace{0.1cm}
  Zhen Dong \hspace{0.1cm}
  Enze Xie \hspace{0.1cm}
  Wojciech Matusik \hspace{0.1cm} 
  Tat-Seng Chua \hspace{0.1cm}
  Daquan Zhou\texorpdfstring{$^{\dagger}$}{}
}

\affiliation[]{PKU, NUS, MIT, UCSB, NVIDIA}
\contribution[*]{Equal Contribution}
\contribution[\dagger]{Corresponding Author}

\newcommand{\answerTODO}[1][]{\textcolor{red}{\bfseries [TODO]}}
\newcommand{\justificationTODO}[1][]{\textcolor{red}{\bfseries [TODO]}}

\usepackage{xcolor}
\usepackage[normalem]{ulem}
\newif\iftrackchanges
\trackchangestrue 

\definecolor{PekingRed}{RGB}{178,31,45}

\abstract{
Embodied foundation models are expected to benefit from data scaling like large language models, but face a much tighter data bottleneck. Teleoperated real-robot trajectories remain the dominant pretraining source due to their precise action supervision and embodiment alignment, yet their scalability is limited by high collection cost, acquisition difficulty, and low behavioral and environmental diversity.
These limitations have sparked interest in egocentric human video as a scalable, substantially lower-cost, and more diverse alternative for embodied model pretraining. However, its effectiveness compared to teleoperated real-robot data remains underexplored.
To address this question, we conduct a systematic study comparing egocentric human video and teleoperated real-robot trajectories as pretraining data sources for embodied foundation models, under fixed post-training and validation protocols.
Surprisingly, we find that egocentric data, when processed through a carefully designed filtering and labeling pipeline, is not merely a viable substitute for model pretraining but can lead to superior performance. 
\textbf{With the same amount of pretraining data, models pretrained on egocentric data achieve a 24\% lower validation loss on real-robot action prediction, as well as 52.5\% and 90\% higher success rates on in-distribution and out-of-distribution real-robot task execution, respectively.}
This finding verifies a scalable paradigm for embodied foundation models: pretrain on egocentric human video to learn diverse world representations, then adapt with a small amount of labeled real-robot data for action-space alignment. 
We hope this study encourages broader exploration of egocentric data and offers guidance for data quality assessment before costly robot data collection.
Code will be released at \url{https://github.com/DAGroup-PKU/HumanNet/}.

\vspace{-6mm}
}

\begin{document}
\maketitle

\section{Introduction}\label{sec:intro}

Foundation models in language and vision have advanced through the joint scaling of data, model size, and compute: heterogeneous, internet-scale corpora provide broad supervision, while empirical scaling laws make the returns from further scaling increasingly predictable~\cite{bai2025qwen3,chen2024expanding,liu2024deepseek,team2025kimi,abouelenin2025phi}. Embodied foundation models, including vision-language-action models (VLAs) and emerging world-action models (WAMs), seek to inherit this scaling recipe, but their data resources follow a different logic~\cite{openx,droid,rt1,rt2}. The dominant source of embodied data is teleoperated real-robot trajectories, which provide valuable action supervision and direct embodiment alignment. Yet, unlike web corpora that can be passively harvested from naturally occurring human behavior, teleoperation data must be actively produced through physical robots, human operators, designed tasks, and controlled environments. This makes it difficult to scale and limits it to a narrow slice of real-world interaction diversity. As a result, the embodied pretraining stage that learns broadly transferable representations before embodiment-specific post-training faces a central tension: the data most aligned with downstream robot policies is poorly to provide the open-world coverage needed for broad generalization.

\begin{figure}[t]
    \centering
    \includegraphics[width=\linewidth]{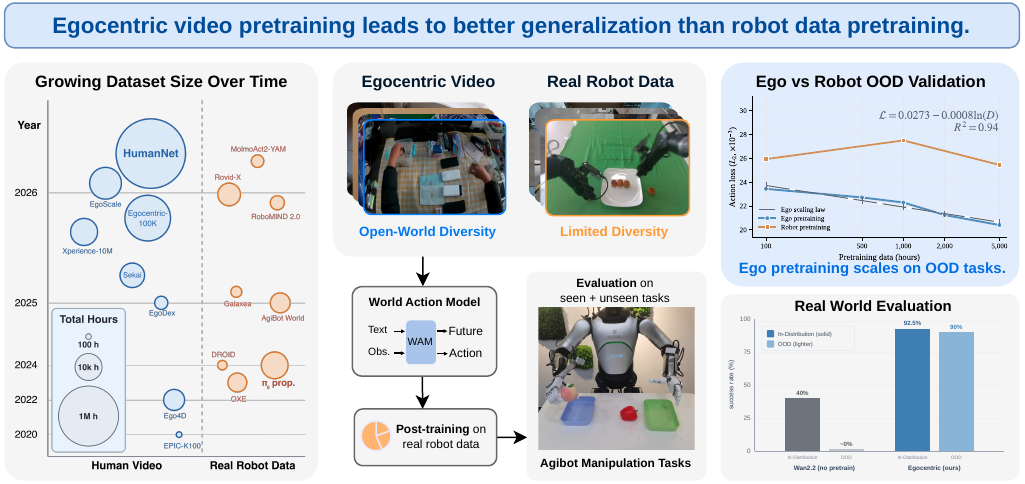}
    \vspace{1mm}
    \caption{\textbf{Egocentric human video leads to stronger generalization than robot data for embodied pretraining.}
\textbf{Left:} Egocentric human video offers massive accessible scale ($\sim10^6$ hours), low collection cost, and open-world diversity, but provides only pseudo action labels from human hand retargeting. Teleoperated robot data provides precise action labels yet is limited in scale ($\sim10^4$ public hours), costly to collect, and limited scene diversity.
\textbf{Middle:} We compare egocentric video pretraining with real robot data pretraining using the same world-action model (WAM). Both pretrained models are post-trained on the same real-robot dataset and evaluated on both seen and unseen AgiBot tasks.
\textbf{Right:} 
On out-of-distribution evaluations, egocentric pretraining exhibits clear scaling behavior, while real-robot pretraining saturates earlier. Real-world rollouts further show that ego-pretrained policies maintain high success rates under unseen-object shifts, whereas the no-pretraining baseline collapses.
}

\label{fig:teasor}
\end{figure}

Egocentric human video offers a natural way to address this coverage gap. 
Captured from a first-person perspective during everyday activity, it exposes models to contact-rich hand-object interactions, tool use, object state changes, and long-horizon behaviors at a scale that teleoperated robotics cannot easily approach.
Recent work has begun to exploit egocentric video for embodied learning, including reusable visual representation learning, human-to-robot motion retargeting, cross-embodiment prior distillation, and downstream manipulation policy learning~\cite{r3m,egomimic,egoscale,egoverse,beingh0,humanego,deng2026rethinking}. These results suggest that human video can provide useful pertraining signals despite lacking robot actions and exact embodiment alignment, but they do not measure whether this coverage advantage is competitive with teleoperated robot data under controlled, matched-scale pretraining. This leaves a basic question open: \emph{how does egocentric human video compare with real-robot data as a pretraining source?}

We answer this question through a controlled, matched-scale comparison and find a clear result: \textbf{egocentric human video pretraining leads to better generalization than real-robot pretraining}. To isolate the effect of the pretraining source, we curate a 5{,}000-hour subset from the egocentric portion of HumanNet~\cite{deng2026humannet}, selected for diversity in objects, skills, and environments, and compare it with an equal-scale multi-embodiment real-robot corpus. Both data sources are used to pretrain the same autoregressive world-action model, built on a Mixture-of-Transformers (MoT) architecture that predicts future video observations and subsequent actions. After pretraining, we post-train each model on the same set of real-robot tasks and evaluate it on two splits: held-out trajectories from the post-training tasks (\emph{Seen}) and held-out tasks not used during post-training (\emph{Unseen}). This design keeps model architecture, data scale, post-training data, and evaluation protocol fixed, making the pretraining source the primary variable.

Our key findings can be summarized as:

\begin{itemize}[leftmargin=*]
    \item \textbf{Egocentric pretraining scales consistently.} As the amount of egocentric pretraining data increases from hunreds of to thousands of hours, downstream validation loss decreases monotonically, indicating that additional egocentric video continues to provide useful embodied pretraining signal.
    \item \textbf{Egocentric pretraining improves downstream generalization over real-robot pretraining.} At matched scale and under the same pretraining--post-training protocol, egocentric pretraining achieves stronger performance than real-robot pretraining, with the largest gains on unseen tasks.
    \item \textbf{The generalization advantage transfers to real-world execution.} On real-robot rollouts,  egocentric pretraining stays robust under distribution shift to unseen objects, while the no-pretraining baseline collapses, showing that the open-world prior holds up beyond validation loss. 
\end{itemize}

\section{Egocentric vs. Real-Robot Data: Coverage, Cost, and Alignment}
\label{sec:data_analysis}

\begin{figure}[t]
    \centering
    \includegraphics[width=0.94\linewidth]{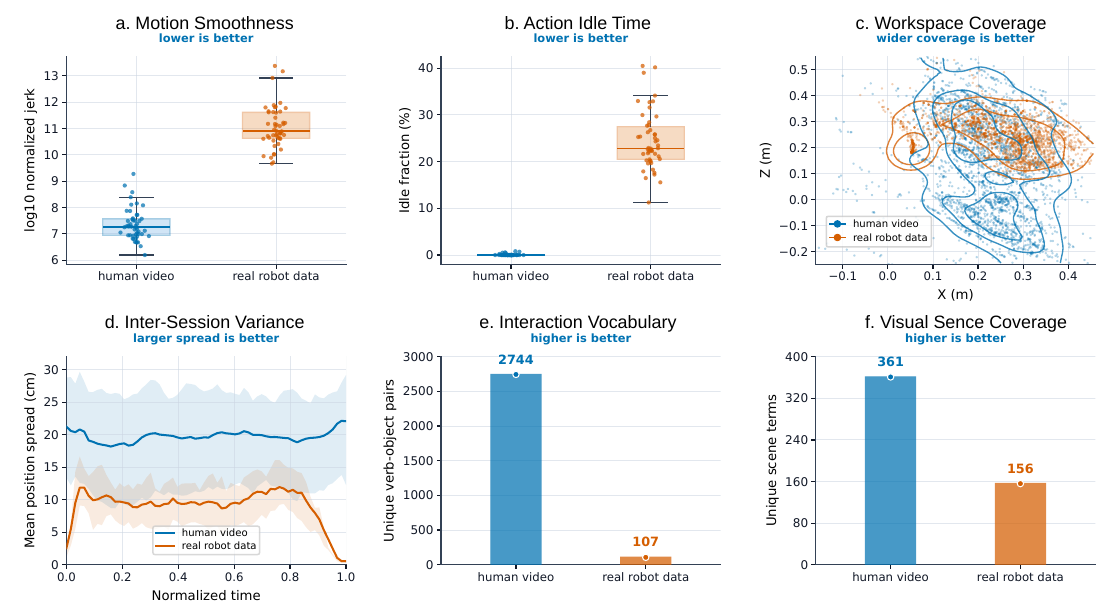}
    \caption{\textbf{Data diversity comparison between our egocentric human video and real-robot data.}
    Following the visualization of HumanEgo~\cite{humanego}, (a--b) motion-quality comparisons show that human trajectories exhibit smoother motion, as reflected by lower normalized jerk, and much less action idle time. (c--d) Spatial and trajectory-diversity comparisons show that human trajectories occupy a broader XZ workspace distribution and maintain larger inter-session positional spread over normalized time. (e--f) Semantic diversity comparisons show that egocentric human video contains a substantially richer interaction vocabulary and broader visual scene coverage than real-robot data. Together, these analyses indicate that egocentric human video provides a cleaner and more diverse pretraining substrate than real-robot teleoperation data at matched sampled duration.}
    \label{fig:data_landscape}
\end{figure}

A practical recipe for embodied models follows the pretraining--post-training paradigm that has been widely validated in language and vision-language modeling: pretraining learns general visual and physical representations from large heterogeneous corpora, while post-training adapts the model to a specific embodiment, camera configuration, and task distribution. Pretraining and post-training therefore emphasize different data properties. Pretraining benefits from \emph{coverage}: broad exposure to scenes, objects, interactions, and behaviors that can support generalizable representations. Post-training, in contrast, places greater emphasis on \emph{alignment}: embodiment-matched observations and actions that adapt the pretrained model to a target embodiment, camera setup, and task distribution. The choice of pretraining data source is thus not about which modality is intrinsically better, but about which supplies coverage at scale. We compare egocentric human video and teleoperated robot data along four axes: accessible scale, collection cost, acquisition difficulty, and diversity (Figure~\ref{fig:data_landscape} and Table~\ref{tab:corpora_scale}).

\subsection{Accessible Scale}
\label{sec:data_scale}
The two data supplies differ by orders of magnitude (Figure~\ref{fig:teasor}, Table~\ref{tab:corpora_scale}). Despite years of community effort, individual robot releases remain in the hundreds to low thousands of hours, and even the most generous aggregations of the entire public supply total only $\sim$$2\times10^4$ hours, comparable to what a single lab holds privately~\cite{beingh05,deng2026rethinking,pi0}. Egocentric video sits an order of magnitude higher: a single release such as Egocentric-100K already exceeds the entire aggregated robot supply several-fold~\cite{egocentric100k}, and frontier systems increasingly pair their robot data with far larger egocentric pools~\cite{beingh05}. HumanNet~\cite{deng2026humannet}, the dataset we build on, curates one million hours of human activity (over 800{,}000 hours of egocentric), from which we draw our 5{,}000-hour pretraining subset. This subset itself is beyond the reach of any open-sourced teleoperation dataset.

\subsection{Collection Cost and Acquisition Difficulty}
\label{sec:data_cost}
The scale gap follows directly from the cost structure. Robot teleoperation requires a robot, a teleoperation rig, a trained operator for the full duration, a staged scene, and ongoing maintenance; even the low-cost ALOHA platform costs $\sim$\$20k per station~\cite{aloha}. As cost is dominated by recurring labor and hardware, total volume scales only linearly with fleet size and budget~\cite{droid,agibotworld}. Egocentric data collection inverts this on every axis: footage is captured passively by an off-the-shelf camera or headset, or harvested from the web at no capture cost at all~\cite{egodex,howto100m,sekai}, which drives the marginal cost of an additional hour far lower. The remaining difficulty shifts from collection to \emph{curation} (filtering, deduplication, privacy review, and pseudo-action labeling via hand-pose retargeting), a one-time, automatable compute cost rather than recurring labor. Human data is also more efficient per unit: minutes of egocentric demonstration can match or exceed far longer teleoperation sessions, with smoother motion and less idle time~\cite{humanego}.

\subsection{Diversity as a Function of Scale}
\label{sec:data_diversity}

Data amount alone is not sufficient for effective embodied pretraining: what matters is whether each additional \emph{marginal} hour exposes the model to new states, motions, interactions, and visual contexts. To compare egocentric human video and real-robot data under a controlled setting, we randomly sample approximately 2-hour subsets from each 5{,}000-hour data pool and compute the statistics shown in Figure~\ref{fig:data_landscape}. Consistent with previous findings in HumanEgo~\cite{humanego}, the sampled egocentric human video also exhibits higher motion quality than real-robot data: human trajectories are smoother, as reflected by lower normalized jerk (Figure~\ref{fig:data_landscape}a), and contain substantially less action idle time (Figure~\ref{fig:data_landscape}b), indicating fewer stationary or uninformative segments. 

Beyond motion quality, the more important question for scaling is whether the data continues to introduce non-redundant experience. At the matched subset scale, egocentric data is substantially more diverse than the collected real-robot data:
\begin{itemize}[leftmargin=*]
    \item \textbf{Workspace coverage}: real-robot motion is concentrated in the compact region reachable from a fixed workstation, whereas egocentric human motion spans a broader XZ workspace induced by unconstrained daily activities (Figure~\ref{fig:data_landscape}c).
    \item \textbf{Inter-session variance}: egocentric trajectories maintain larger positional variation across demonstrations, indicating that different demonstrations occupy more varied spatial configurations rather than repeatedly traversing a narrow motion manifold. Following HumanEgo~\cite{humanego}, we quantify this effect by measuring positional spread across demonstrations over normalized time (Figure~\ref{fig:data_landscape}d).
    \item \textbf{Interaction vocabulary}: we count the set of unique verb-object pairs extracted from task descriptions. Robot data is constrained by scripted collection tasks and therefore repeats limited interactions, while egocentric video exhibits a long-tailed, open-vocabulary distribution of human-object interactions (Figure~\ref{fig:data_landscape}e).
    \item \textbf{Visual scene coverage}: we measure the diversity of scene semantics from caption-derived scene terms. Real-robot data saturates quickly because it is collected in bounded environments, while egocentric video covers a wider range of homes, workshops, kitchens, outdoor settings, objects, and surfaces (Figure~\ref{fig:data_landscape}f).
\end{itemize}

Together, these axes show that real-robot data saturation is structural, reflecting bounded environments, fixed workspace coverage, and scripted tasks, rather than an artifact of a particular collection effort. In contrast, egocentric human video is both individually informative and collectively less redundant: at the same sampled duration, it exposes the model to broader motion, spatial, interaction, and visual variation.

\begin{table}[t]
\caption{Landmark publicly released datasets with reported durations, sorted by scale. $\dagger$ marks datasets that aggregate other releases; $\ddagger$ marks datasets containing simulated trajectories.}
\label{tab:corpora_scale}
\centering
\setlength{\tabcolsep}{5pt}
\resizebox{\linewidth}{!}{%
\begin{tabular}{lrll}
\toprule
Corpus & Hours & Acquisition & Coverage \\
\midrule
\rowcolor{RowRed}
\multicolumn{4}{l}{\textit{Egocentric human video}} \\
EgoDex~\cite{egodex} & 829 & Consumer headset, native hand pose & 194 tabletop manipulation tasks \\
Ego4D~\cite{ego4d} & 3{,}670 & 931 camera wearers, daily life & 74 locations, 9 countries \\
Sekai~\cite{sekai} & 5{,}000+ & Web harvest (egocentric POV) & Walking/exploration, global \\
Xperience-10M~\cite{xperience_10m} & 10{,}000 & Wearable capture, 10M interactions & Open-world daily experience \\
EgoScale~\cite{egoscale}$^\dagger$ & 20{,}854 & Aggregation + hand retargeting & Dexterous manipulation \\
Egocentric-100K~\cite{egocentric100k} & 100{,}405 & 14{,}228 workers, head-mounted glasses & Industrial / factory operations \\
\midrule
\rowcolor{RowRed}
\multicolumn{4}{l}{\textit{Teleoperated real-robot}} \\
DROID~\cite{droid} & 350 & 50 collectors, 12 months & 564 scenes, 84 tasks \\
Galaxea Open-World~\cite{galaxea_g0} & 500 & Single embodiment, in-the-wild & Homes, kitchens, retail, offices \\
MolmoAct2 BimanualYAM~\cite{molmoact2} & 720+ & Bimanual YAM arms & Largest open bimanual release \\
RoboMIND 2.0~\cite{robomind2} & 1{,}000+ & Bimanual mobile teleop, 310K+ trajs & Bimanual coordination tasks \\
Open X-Embodiment~\cite{openx}$^\dagger$ & $\sim$2{,}000--3{,}000 & Pooling of 60 datasets, 1M+ trajs & 22 embodiments, 527 skills \\
AgiBot World~\cite{agibotworld} & 2{,}976 & 100-robot fleet, 1M+ trajs & 217 tasks, 106 scenes, 5 domains \\
RoVid-X~\cite{deng2026rethinking}$^{\dagger\ddagger}$ & $10{,}000$+ & Open-source aggregation, 4M robot videos & 1{,}300+ fine-grained robot skills \\
{Being H-0.5}~\cite{beingh05}$^{\dagger\ddagger}$ & $\sim$35{,}000 & OXE + AgiBot + RoboMIND + RoboCOIN + \ldots & 30 embodiments, incl.\ sim \\
\midrule
\rowcolor{RowRed}
\multicolumn{4}{l}{\textit{Reference points}} \\
$\pi_0$ corpus~\cite{pi0} & $>$10{,}000 & Internal fleet, 7 platforms & Proprietary, inaccessible \\
HumanNet~\cite{deng2026humannet} & 1{,}000{,}000 & Web curation, egocentric + third-person & Open-world, long-tail interaction \\
\bottomrule
\end{tabular}%
}
\end{table}

\medskip
\noindent\textbf{A division of labor.}
The analysis yields a clear division of labor in embodied training. Egocentric video dominates the axes that pretraining rewards, including scale, marginal cost, motion diversity, interaction diversity, and scene diversity, while its main weakness, the embodiment gap, is precisely what post-training can correct with a smaller amount of kinematically aligned robot data. Teleoperation data is strongest on the axis post-training needs, namely embodiment alignment, and weakest on the axes pretraining needs. \textbf{The open question is empirical: at matched scale, does the diversity advantage of egocentric pretraining outweigh the kinematic alignment advantage of robot pretraining in the downstream policy?} The rest of this paper answers in the affirmative through a controlled comparison.

\section{Embodied Pretraining with Egocentric Human Video}
\label{sec:method}
We study egocentric pretraining with an autoregressive world action model that unifies video dynamics prediction and action inference through a Mix-of-Transformers (MoT) architecture. 
Specifically, the video expert is initialized from \emph{Wan 2.2}, while the action expert is initialized via interpolation. 
To isolate the effect of the pretraining substrate, we compare the post-training  performance of models pretrained on egocentric human video versus real-robot data. Throughout these comparisons, we rigorously hold the post-training data, compute budget, and evaluation protocol fixed.

The splits are designed around one principle: introduce a controlled distributional shift at each stage so that downstream performance isolates the quality of the pretraining substrate. We pretrain separately on egocentric and real-robot data, post-train on a real-robot dataset disjoint from both, and evaluate on two held-out splits that respectively measure in-distribution robustness and out-of-distribution extrapolation.

\textbf{Stage 1: Pretraining data.} We construct two pretraining sets matched at 5{,}000 hours but differing in collection:
\begin{description}[leftmargin=0pt, labelindent=0pt, itemsep=2pt, parsep=0pt]
    \item[Egocentric.] Curated from the egocentric portion of HumanNet~\cite{deng2026humannet}; per-clip end-effector poses and gripper states are estimated from retargeted hand-pose signals as pseudo-action labels, placing it in the same action space as the robot data. It offers open-world coverage of backgrounds, objects, and interaction skills.
    \item[Real-robot.] 
    Multi-embodiment trajectories with precise end-effector poses and gripper states, aggregated from various real-robot datasets. Although its environmental diversity and task scenarios remain bounded compared to the egocentric counterpart, it provides kinematically aligned embodied priors.
\end{description}

\textbf{Stage 2: Post-training Data.}
\label{sec:posttrain_data}
For post-training, we curate a real-robot dataset from AgiBot World~\cite{agibotworld}, selecting 15 manipulation tasks with 100 expert demonstrations per task, resulting in 1{,}500 trajectories in total. Compared with the pretraining robot data, this post-training set contains more diverse backgrounds and object instances, providing richer downstream manipulation scenarios for adapting the pretrained policy.

\textbf{Evaluation Protocol.}
\label{sec:eval_protocol}
We evaluate models using validation flow-matching action loss on held-out Stage-2 robot data. We report results on two splits. The \emph{Seen} split holds out trajectories from the 15 post-training tasks, where the task semantics are observed during post-training but the evaluation trajectories involve unseen object instances and variations, measuring robustness within the post-training task distribution. The \emph{Unseen} split comprises 25 tasks that are not included in Stage-2 post-training, and serves as our primary evaluation for out-of-distribution generalization. This split tests whether the open-world prior learned from egocentric pretraining can better transfer to novel manipulation scenarios than a kinematically aligned prior learned from real-robot pretraining with relatively limited scene and object coverage.

\begin{figure}[t]
\centering
\includegraphics[width=0.95\linewidth]{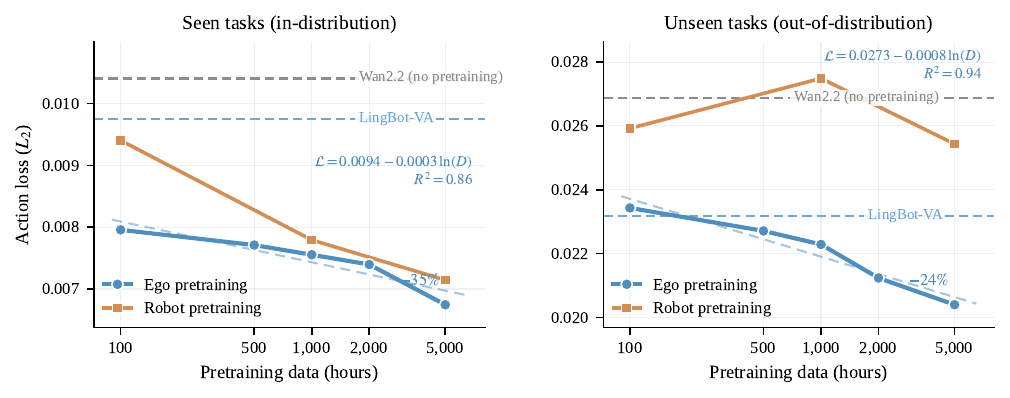}
\caption{\textbf{Egocentric pretraining scales log-linearly while robot pretraining stalls on unseen tasks.} Best post-training action loss as a function of pretraining hours on Seen and Unseen tasks. The blue curve shows ego pretraining, and the light orange curve overlays robot pretraining at matched scales. Dashed gray and blue lines denote the Wan2.2 \cite{wan2025} (no pretraining) and LingBot-VA~\cite{li2026causal} references. }
\label{fig:ego_scaling_robot}
\vspace{-5pt}
\end{figure}

\section{Experimental Results and Analysis}
\label{sec:experiments}

In this section, we conduct empirical studies around two questions that together characterize the value of egocentric pretraining for embodied foundation models.

\begin{figure}[t]
\centering
\includegraphics[width=\linewidth]{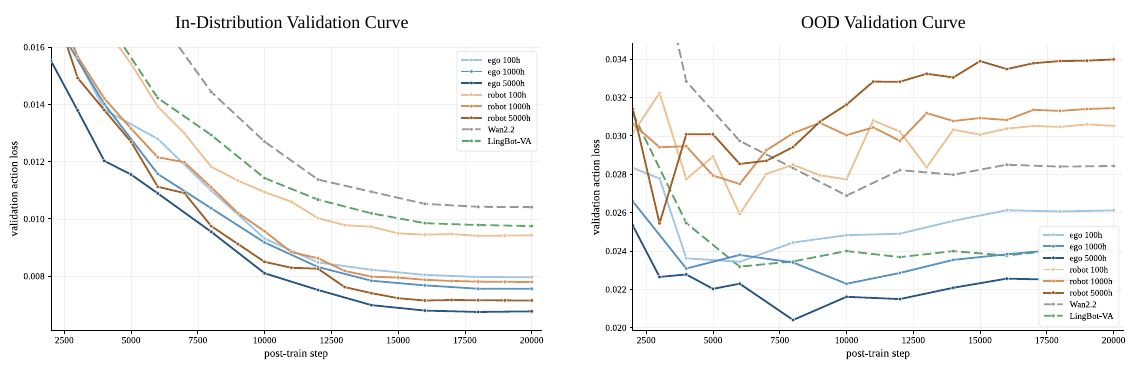}
\caption{Complete validation action loss curves during post-training, for egocentric and real-robot pretraining at 100, 1{,}000, and 5{,}000 hours together with the Wan2.2 and LingBot-VA baselines. The left panel reports seen tasks (in-distribution), and the right panel reports unseen tasks (out-of-distribution).}
\label{fig:trainloss_comparison}
\end{figure}

\begin{itemize}[leftmargin=*]
\item \textbf{Q1 (Ego Pretrain Scaling):} Does egocentric pretraining exhibit scaling behavior in robot post-training?

\item \textbf{Q2 (Ego vs. Robot Pretrain):} How does egocentric pretraining perform compared to real-robot pretraining?

\end{itemize}
We follow the training and evaluation protocol described in Section~\ref{sec:method}. We evaluate checkpoints throughout post-training and report the minimum validation loss on both in-distribution and out-of-distribution tasks. We compare our method with two baselines. The first is Wan2.2 without embodied pretraining. The second is LingBot-VA, which fine-tunes Wan2.2 on 20k hours of real-robot data and serves as a strong embodied-pretrained baseline. The complete validation action loss curves are shown in Fig.~\ref{fig:trainloss_comparison}, and the analysis below summarizes each curve by its lowest loss.

\subsection{Egocentric Pretraining Scales with Data}
\label{sec:scaling}
As egocentric pretraining scales from 100 to 5{,}000 hours, the best post-training action loss decreases monotonically in both evaluation settings (Fig.~\ref{fig:ego_scaling_robot}). Specifically, the loss drops from 0.0080 to 0.0067 on seen tasks and from 0.0234 to 0.0204 on unseen tasks, reaching values $35\%$ and $24\%$ lower than the Wan2.2 baseline without pretraining. Across this range, the trend is well captured by a log-linear scaling law, $\mathcal{L} = a - b\ln(D)$, with $R^2 = 0.86$ for seen tasks and $R^2 = 0.94$ for unseen tasks. The fitted slope remains clearly negative up to 5{,}000 hours, suggesting that egocentric pretraining has not yet saturated and that further gains may be possible as the data scale increases.

We attribute this scaling behavior to the diversity of the egocentric data. Because our egocentric subsets are curated for broad coverage, increasing the number of hours does not simply repeat redundant trajectories, but introduces a wider range of manipulated objects, manipulation skills, and physical environments. The growing diversity benefits the two evaluation settings in complementary ways. For unseen tasks, out-of-distribution generalization depends more on the breadth of behavioral space observed during pretraining, and broader coverage therefore leads to lower loss on tasks held out from post-training. For seen tasks, the same diversity improves the pretrained representation. When the pretraining data are narrow, the model fits a limited set of human activities and transfers a correspondingly specialized initialization to post-training. In contrast, a diverse corpus encourages a more general and less task-specific representation, which provides a stronger starting point for adaptation to the target robot tasks.

\subsection{Egocentric Pretraining Generalizes Better than Robot Pretraining}
We compare egocentric and real-robot pretraining at matched data scales (Fig.~\ref{fig:ego_scaling_robot}). The two data sources exhibit distinct scaling behaviors across the two evaluation settings. On seen tasks, real-robot pretraining improves steadily with data scale and remains close to egocentric pretraining, reaching a loss of 0.0071 at 5{,}000 hours, compared with 0.0067 for egocentric pretraining. On unseen tasks, however, scaling real-robot data produces no consistent improvement: its loss remains near 0.025 across all scales and reaches 0.0254 at 5{,}000 hours, substantially higher than that of egocentric pretraining. In contrast, egocentric pretraining continues to improve on unseen tasks as the data scale increases, reaching 0.0204 at 5{,}000 hours and achieving a roughly 20\% lower loss than real-robot pretraining at the same scale. These results indicate that the two sources are comparably effective for in-distribution transfer, whereas only egocentric data scales toward stronger out-of-distribution generalization. 

The difference derives from two factors:

\begin{itemize}[leftmargin=*]
\item \textbf{Diversity and information density.} As analyzed in Section~\ref{sec:data_analysis}, egocentric data is intrinsically more diverse and covers a far wider range of tasks, objects, and backgrounds than teleoperated robot data, which is collected in a bounded set of laboratory setups. Matching the two sources by hours further understates this advantage, since an hour of egocentric video contains far more and cleaner trajectories than an hour of robot teleoperation. In our 100-hour recipe, for instance, the egocentric data comprises roughly 45{,}000 trajectories, whereas the real-robot data contains only about 8{,}000, as teleoperation is slowed by long idle intervals and the comparatively slow motion of the robot arm. Each hour of teleoperated data therefore carries substantially less information and offers little advantage as a pretraining source.

\item \textbf{Limited generalization to unseen skills.} We keep the real-robot pretraining tasks disjoint from both the post-training and the out-of-distribution evaluation tasks, and preserve this isolation as the real-robot corpus is scaled up. This control rules out any shortcut in which scaled robot data overlaps with the evaluation tasks, so the measured behavior reflects genuine generalization rather than leakage. However, scaling real-robot pretraining does not yield the generalization, as its loss on Unseen tasks stays nearly flat while the data grows, indicating that manipulation tasks collected in a constrained laboratory setting transfer poorly to genuinely unseen tasks.
\end{itemize}

\begin{figure}[t]
\centering
\includegraphics[width=0.95\linewidth]{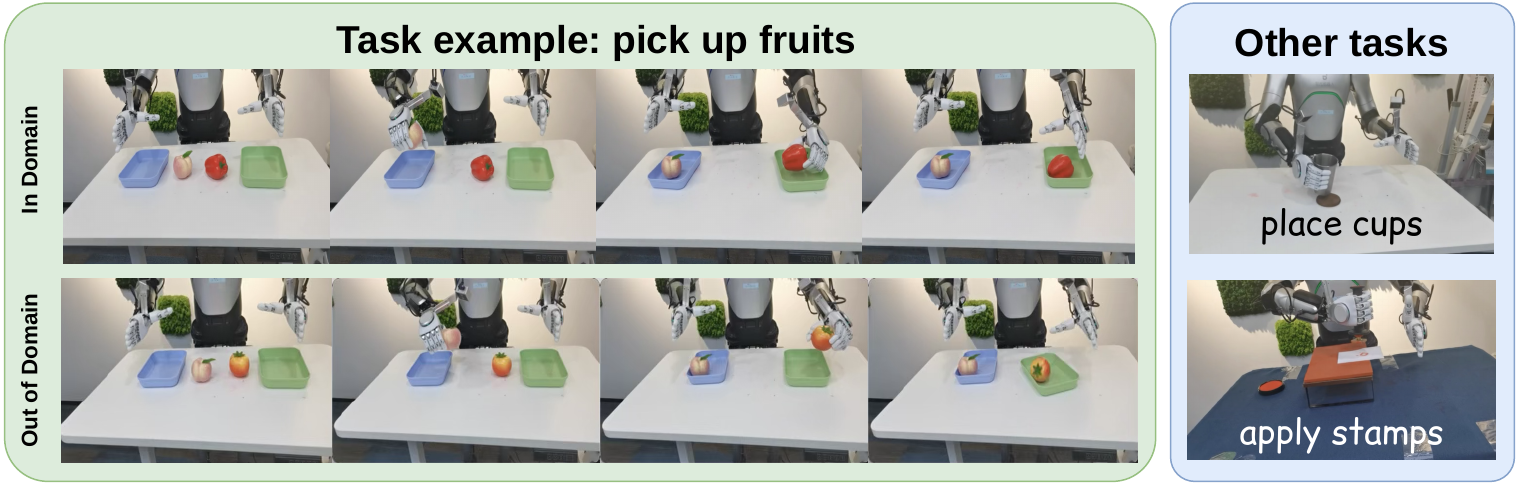}
\caption{Real-world rollouts on three tasks (place cup, pick up fruits, stamp), each under in-distribution and ood settings. The egocentric-pretrained model succeeds under distribution shift, whereas the baseline fails.}
\label{fig:robotexp}
\end{figure}

\subsection{Real-World Robot Experiments}
\begin{figure}[t]
\centering
\includegraphics[width=0.9\linewidth]{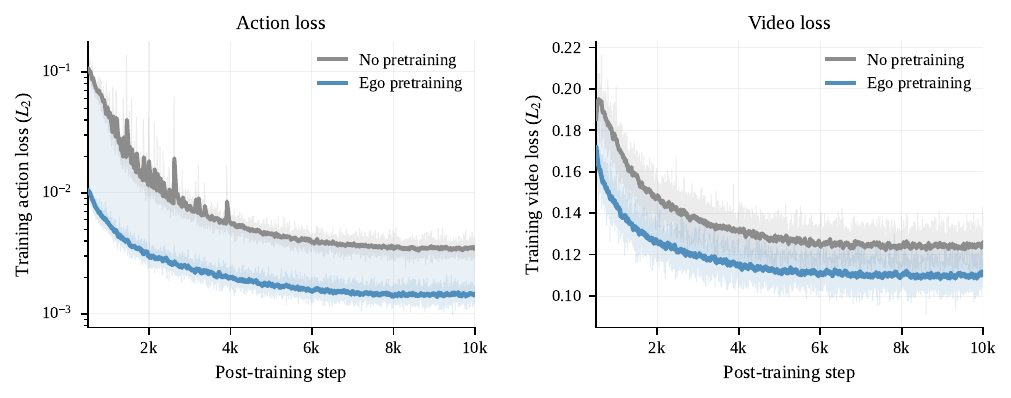}
\caption{Post-training action and video loss on the fruit-and-vegetable sorting task. Egocentric pretraining stays below the no-pretraining baseline throughout, converging to about $2.4\times$ lower action loss as well as lower video loss.}
\label{fig:trainloss}
\end{figure}

We validate our findings with real-robot rollouts on an AgiBot bimanual platform across three manipulation tasks: placing a cup on a coaster, sorting fruits and vegetables, and stamping. For each task we evaluate an in-distribution setting, whose objects are seen during post-training, and an out-of-distribution setting with previously unseen object instances. We compare the egocentric-pretrained model against the Wan 2.2 baseline under identical post-training. The advantage of egocentric pretraining is already visible during post-training. On the fruit-and-vegetable sorting task (Fig.~\ref{fig:trainloss}), the egocentric-pretrained initialization starts with a substantially lower loss and converges to a value approximately $2.4\times$ lower than that of the no-pretraining baseline. This result indicates that the pretrained prior makes the downstream task easier to fit and is consistent with the observed gap in the following real-world rollouts.

\begin{table}[h]
\centering
\caption{Real-robot performance on AgiBot bimanual platform. Success rate is averaged over tasks. Egocentric pretraining transfers to out-of-distribution objects with little degradation, while the baseline collapses.}
\label{tab:realrobot}
\begin{tabular}{lcc}
\toprule
Pretraining & In-distribution & Out-of-distribution \\
\midrule
Wan2.2 (baseline) & 40.0\% & 0.0\% \\
Egocentric (ours) & \textbf{92.5\%} & \textbf{90.0\%} \\
\bottomrule
\end{tabular}
\end{table}

As summarized in Table~\ref{tab:realrobot}, the egocentric-pretrained model attains a $92.5\%$ in-distribution success rate and retains $90.0\%$ under distribution shift with a drop of only $2.5$ points. The baseline, by contrast, reaches $40.0\%$ in-distribution and degrades to $0\%$ on ood trials, a collapse of $40$ points. The degradation is small for egocentric pretraining but catastrophic for the baseline, indicating that the open-world prior learned from human video transfers, whereas a policy lacking this prior overfits to the narrow visual distribution of its post-training data and fails to generalize.

\section{Related Work}
\label{sec:related}
   
\textbf{Scaling robot learning with real-robot data.} Teleoperated data is the dominant substrate for embodied foundation models. Open X-Embodiment federated over one million trajectories across 22 embodiments, showing cross-embodiment mixing improves robustness and zero-shot transfer~\cite{openx}. RT-1/RT-2 established that scaling robot-hours and fine-tuning a web-pretrained VLM backbone yield semantic generalization~\cite{rt1,rt2}. DROID~\cite{droid}, $\pi_0$~\cite{pi0}, GR00T~\cite{gr00t}, and AgiBot World~\cite{agibotworld} pushed scale, diversity, and new embodiments. This data is kinematically aligned with the deployment policy, but its collection cost bounds supply to current robots, operators, and labs, yielding homogeneous backgrounds, objects, and interactions, as well as brittle out-of-distribution behavior.

\textbf{Egocentric data for robot pretraining.} Egocentric corpora such as EPIC-KITCHENS~\cite{egokitchens}, Ego4D~\cite{ego4d}, and Ego-Exo4D~\cite{egoexo4d} have long supported recognition and representation learning, with embodied transfer as an indirect application. A recent line targets egocentric video directly as a pretraining substrate: EgoMimic co-trains human and robot data~\cite{egomimic}; EgoScale scales to $20{,}000$+ hours with retargeted hand and wrist motion and reports a log-linear scaling law on dexterous manipulation~\cite{egoscale}; Being-H0 distills a hand-trajectory prior for humanoids~\cite{beingh0}; and HumanEgo shows minutes of egocentric demonstration can replace far longer teleoperation~\cite{humanego}. Yet no prior work compares egocentric and real-robot pretraining \emph{head-to-head at matched scale} on a controlled post-training benchmark. We fill this gap, showing egocentric pretraining not only substitutes for but surpasses real-robot pretraining, with the largest gains on OOD generalization, and we establish the first such scaling curves for an autoregressive world-action model.

\textbf{Architectures for embodied intelligence.} Embodied policy learning is organized around two architectural families. Vision-language-action (VLA) models add an action head to a pretrained vision-language backbone, as in RT-2~\cite{rt2}, $\pi_0$~\cite{pi0}, GR00T~\cite{gr00t}, OpenVLA~\cite{kim24openvla}, RDT~\cite{liu2024rdt}, and LingBot-VLA~\cite{lingbotvla}. World-Action Models (WAMs) instead model future states and actions jointly, using video generation as a dense representation of how the world evolves under control. DreamZero~\cite{ye2026worldactionmodelszeroshot} jointly denoises future video and actions in a single diffusion process, while LingBot-VA~\cite{li2026causal} predicts them sequentially, first generating future video and then decoding the corresponding actions via causal autoregression. Since the imagine-then-execute paradigm incurs heavy test-time latency from iterative video rollout, Fast-WAM~\cite{yuan2026fastwamworldactionmodels} retains video co-training but skips future generation at inference, showing that the benefit of video modeling lies mainly in shaping representations during training rather than imagining futures at test time. The two families exploit visual supervision differently. Our experiments focus on the world-action model, where video generation provides a dense learning signal alongside action prediction; we leave a parallel study of VLA models to future work.

\section{Conclusion}
\label{sec:conclusion}
We presented a controlled, matched-scale comparison of egocentric human video and real-robot data as pretraining data source for embodied foundation models. From the perspective of pretraining data, egocentric video leads on the axes that pretraining rewards, namely scale, cost, and diversity, while its embodiment gap is the part post-training is meant to close. Empirically, under an identical post-training and evaluation protocol, egocentric pretraining scales with data and surpasses real-robot pretraining, with the largest gains on out-of-distribution generalization. We view these results as encouraging, but still preliminary.

\textbf{Future Work.} We have a few ongoing explorations. Due to the limited availability of real-robot data, the current training scheme is limited to 5000 hours. As more real-robot data become available, we plan to scale the egocentric corpus and pretraining budget substantially further. Besides, current evaluations are mainly based on world action models with WAN2.2 serving as the backbone of the video generation model. 
Currently, we are also evaluating the scaling behavior of egocentric data based on vision-language-action models (VLAs) across a broader range of robot embodiments, testing whether the advantage we observe persists at foundation-model scale and beyond the world-action model studied here.

\clearpage

\bibliographystyle{plainnat}
\setlength{\bibhang}{0pt}
\setlength\bibindent{0pt}
\bibliography{main}

@article{team2025kimi,
  title={Kimi-vl technical report},
  author={Team, Kimi and Du, Angang and Yin, Bohong and Xing, Bowei and Qu, Bowen and Wang, Bowen and Chen, Cheng and Zhang, Chenlin and Du, Chenzhuang and Wei, Chu and others},
  journal={arXiv preprint arXiv:2504.07491},
  year={2025}
}

@article{abouelenin2025phi,
  title={Phi-4-mini technical report: Compact yet powerful multimodal language models via mixture-of-loras},
  author={Abouelenin, Abdelrahman and Ashfaq, Atabak and Atkinson, Adam and Awadalla, Hany and Bach, Nguyen and Bao, Jianmin and Benhaim, Alon and Cai, Martin and Chaudhary, Vishrav and Chen, Congcong and others},
  journal={arXiv preprint arXiv:2503.01743},
  year={2025}
}

@article{liu2024deepseek,
  title={Deepseek-v3 technical report},
  author={Liu, Aixin and Feng, Bei and Xue, Bing and Wang, Bingxuan and Wu, Bochao and Lu, Chengda and Zhao, Chenggang and Deng, Chengqi and Zhang, Chenyu and Ruan, Chong and others},
  journal={arXiv preprint arXiv:2412.19437},
  year={2024}
}

@article{chen2024expanding,
  title={Expanding performance boundaries of open-source multimodal models with model, data, and test-time scaling},
  author={Chen, Zhe and Wang, Weiyun and Cao, Yue and Liu, Yangzhou and Gao, Zhangwei and Cui, Erfei and Zhu, Jinguo and Ye, Shenglong and Tian, Hao and Liu, Zhaoyang and others},
  journal={arXiv preprint arXiv:2412.05271},
  year={2024}
}

@article{bai2025qwen3,
  title={Qwen3-vl technical report},
  author={Bai, Shuai and Cai, Yuxuan and Chen, Ruizhe and Chen, Keqin and Chen, Xionghui and Cheng, Zesen and Deng, Lianghao and Ding, Wei and Gao, Chang and Ge, Chunjiang and others},
  journal={arXiv preprint arXiv:2511.21631},
  year={2025}
}

@dataset{xperience_10m,
  title={Xperience-10M: A Large-Scale Egocentric Multimodal Dataset with Structured 3D/4D Annotations},
  author={Ropedia},
  year={2026},
  publisher={Hugging Face},
  note={Dataset}
}

@misc{yuan2026fastwamworldactionmodels,
      title={Fast-WAM: Do World Action Models Need Test-time Future Imagination?}, 
      author={Tianyuan Yuan and Zibin Dong and Yicheng Liu and Hang Zhao},
      year={2026},
      eprint={2603.16666},
      archivePrefix={arXiv},
      primaryClass={cs.CV},
      url={https://arxiv.org/abs/2603.16666}, 
}

@article{li2026causal,
  title={Causal World Modeling for Robot Control},
  author={Li, Lin and Zhang, Qihang and Luo, Yiming and Yang, Shuai and Wang, Ruilin and Han, Fei and Yu, Mingrui and Gao, Zelin and Xue, Nan and Zhu, Xing and others},
  journal={arXiv preprint arXiv:2601.21998},
  year={2026}
}

@article{rt1,
  title={Rt-1: Robotics transformer for real-world control at scale},
  author={Brohan, Anthony and Brown, Noah and Carbajal, Justice and Chebotar, Yevgen and Dabis, Joseph and Finn, Chelsea and Gopalakrishnan, Keerthana and Hausman, Karol and Herzog, Alex and Hsu, Jasmine and others},
  journal={arXiv preprint arXiv:2212.06817},
  year={2022}
}

@inproceedings{rt2,
  title={Rt-2: Vision-language-action models transfer web knowledge to robotic control},
  author={Zitkovich, Brianna and Yu, Tianhe and Xu, Sichun and Xu, Peng and Xiao, Ted and Xia, Fei and Wu, Jialin and Wohlhart, Paul and Welker, Stefan and Wahid, Ayzaan and others},
  booktitle={Conference on Robot Learning},
  pages={2165--2183},
  year={2023},
  organization={PMLR}
}

@misc{droid,
  title         = {DROID: A Large-Scale In-The-Wild Robot Manipulation Dataset},
  author        = {Khazatsky, Alexander and others},
  year          = {2025},
  eprint        = {2403.12945},
  archivePrefix = {arXiv},
  primaryClass  = {cs.RO},
  url           = {https://arxiv.org/abs/2403.12945},
}

@article{deng2026humannet,
  title={HumanNet: Scaling Human-centric Video Learning to One Million Hours},
  author={Deng, Yufan and Zhou, Daquan},
  journal={arXiv preprint arXiv:2605.06747},
  year={2026}
}

@misc{howto100m,
      title={HowTo100M: Learning a Text-Video Embedding by Watching Hundred Million Narrated Video Clips}, 
      author={Antoine Miech and Dimitri Zhukov and Jean-Baptiste Alayrac and Makarand Tapaswi and Ivan Laptev and Josef Sivic},
      year={2019},
      eprint={1906.03327},
      archivePrefix={arXiv},
      primaryClass={cs.CV},
      url={https://arxiv.org/abs/1906.03327}, 
}

@misc{egodex,
      title={EgoDex: Learning Dexterous Manipulation from Large-Scale Egocentric Video}, 
      author={Ryan Hoque and Peide Huang and David J. Yoon and Mouli Sivapurapu and Jian Zhang},
      year={2026},
      eprint={2505.11709},
      archivePrefix={arXiv},
      primaryClass={cs.CV},
      url={https://arxiv.org/abs/2505.11709}, 
}

@misc{egoverse,
  title         = {EgoVerse: An Egocentric Human Dataset for Robot Learning from Around the World},
  author        = {Punamiya, Ryan and others},
  year          = {2026},
  eprint        = {2604.07607},
  archivePrefix = {arXiv},
  primaryClass  = {cs.RO},
  url           = {https://arxiv.org/abs/2604.07607},
}

@misc{egoscale,
      title={EgoScale: Scaling Dexterous Manipulation with Diverse Egocentric Human Data}, 
      author={Ruijie Zheng and Dantong Niu and Yuqi Xie and Jing Wang and Mengda Xu and Yunfan Jiang and Fernando Castañeda and Fengyuan Hu and You Liang Tan and Letian Fu and Trevor Darrell and Furong Huang and Yuke Zhu and Danfei Xu and Linxi Fan},
      year={2026},
      eprint={2602.16710},
      archivePrefix={arXiv},
      primaryClass={cs.RO},
      url={https://arxiv.org/abs/2602.16710}, 
}

@inproceedings{ego4d,
  title     = {{Ego4D}: Around the World in 3,000 Hours of Egocentric Video},
  author    = {Grauman, Kristen and Westbury, Andrew and Byrne, Eugene and Chavis, Zachary and others},
  booktitle = {Proceedings of the IEEE/CVF Conference on Computer Vision and Pattern Recognition},
  year      = {2022},
}

@article{egokitchens,
  title={Rescaling egocentric vision: Collection, pipeline and challenges for epic-kitchens-100},
  author={Damen, Dima and Doughty, Hazel and Farinella, Giovanni Maria and Furnari, Antonino and Kazakos, Evangelos and Ma, Jian and Moltisanti, Davide and Munro, Jonathan and Perrett, Toby and Price, Will and others},
  journal={International Journal of Computer Vision},
  volume={130},
  number={1},
  pages={33--55},
  year={2022},
  publisher={Springer}
}

@misc{openx,
  title         = {{Open X-Embodiment}: Robotic Learning Datasets and {RT-X} Models},
  author        = {{Open X-Embodiment Collaboration} and others},
  year          = {2025},
  eprint        = {2310.08864},
  archivePrefix = {arXiv},
  primaryClass  = {cs.RO},
  url           = {https://arxiv.org/abs/2310.08864},
}

@misc{r3m,
      title={R3M: A Universal Visual Representation for Robot Manipulation}, 
      author={Suraj Nair and Aravind Rajeswaran and Vikash Kumar and Chelsea Finn and Abhinav Gupta},
      year={2022},
      eprint={2203.12601},
      archivePrefix={arXiv},
      primaryClass={cs.RO},
      url={https://arxiv.org/abs/2203.12601}, 
}

@misc{beingh05,
      title={Being-H0.5: Scaling Human-Centric Robot Learning for Cross-Embodiment Generalization}, 
      author={Hao Luo and Ye Wang and Wanpeng Zhang and Sipeng Zheng and Ziheng Xi and Chaoyi Xu and Haiweng Xu and Haoqi Yuan and Chi Zhang and Yiqing Wang and Yicheng Feng and Zongqing Lu},
      year={2026},
      eprint={2601.12993},
      archivePrefix={arXiv},
      primaryClass={cs.RO},
      url={https://arxiv.org/abs/2601.12993}, 
}

@misc{beingh0,
      title={Being-H0: Vision-Language-Action Pretraining from Large-Scale Human Videos}, 
      author={Hao Luo and Yicheng Feng and Wanpeng Zhang and Sipeng Zheng and Ye Wang and Haoqi Yuan and Jiazheng Liu and Chaoyi Xu and Qin Jin and Zongqing Lu},
      year={2025},
      eprint={2507.15597},
      archivePrefix={arXiv},
      primaryClass={cs.CV},
      url={https://arxiv.org/abs/2507.15597}, 
}

@misc{egoexo4d,
    title         = {{Ego-Exo4D}: Understanding Skilled Human Activity from First- and Third-Person Perspectives},
    author        = {Grauman, Kristen and Westbury, Andrew and Torresani, Lorenzo and Kitani, Kris and others},
    year          = {2024},
    eprint        = {2311.18259},
    archivePrefix = {arXiv},
    primaryClass  = {cs.CV},
    url           = {https://arxiv.org/abs/2311.18259},
}

@misc{egomimic,
      title={EgoMimic: Scaling Imitation Learning via Egocentric Video}, 
      author={Simar Kareer and Dhruv Patel and Ryan Punamiya and Pranay Mathur and Shuo Cheng and Chen Wang and Judy Hoffman and Danfei Xu},
      year={2024},
      eprint={2410.24221},
      archivePrefix={arXiv},
      primaryClass={cs.RO},
      url={https://arxiv.org/abs/2410.24221}, 
}

@article{deng2026rethinking,
  title={Rethinking Video Generation Model for the Embodied World},
  author={Deng, Yufan and Pan, Zilin and Zhang, Hongyu and Li, Xiaojie and Hu, Ruoqing and Ding, Yufei and Zou, Yiming and Zeng, Yan and Zhou, Daquan},
  journal={arXiv preprint arXiv:2601.15282},
  year={2026}
}

@misc{gr00t,
    title         = {{GR00T N1}: An Open Foundation Model for Generalist Humanoid Robots},
    author        = {{NVIDIA} and others},
    year          = {2025},
    eprint        = {2503.14734},
    archivePrefix = {arXiv},
    primaryClass  = {cs.RO},
    url           = {https://arxiv.org/abs/2503.14734},
}

@article{lingbotvla,
  title={A Pragmatic VLA Foundation Model},
  author={Wu, Wei and Lu, Fan and Wang, Yunnan and Yang, Shuai and Liu, Shi and Wang, Fangjing and Zhu, Qian and Sun, He and Wang, Yong and Ma, Shuailei and others},
  journal={arXiv preprint arXiv:2601.18692},
  year={2026}
}

@article{agibotworld,
  title={AgiBot World Colosseo: A Large-scale Manipulation Platform for Scalable and Intelligent Embodied Systems},
  author={{AgiBot World Contributors}},
  journal={arXiv preprint arXiv:2503.06669},
  year={2025}
}

@article{pi0,
  title={$\pi_0$: A Vision-Language-Action Flow Model for General Robot Control},
  author={Black, Kevin and Brown, Noah and Driess, Danny and Esmail, Adnan and Equi, Michael and Finn, Chelsea and Fusai, Niccolo and Groom, Lachy and Hausman, Karol and Ichter, Brian and others},
  journal={arXiv preprint arXiv:2410.24164},
  year={2024}
}

@article{humanego,
  title={HumanEgo: Zero-Shot Robot Learning from Minutes of Human Egocentric Videos},
  author={Wang, Zhi and He, Botao and Yu, Kelin and Lee, Seungjae and Gao, Ruohan and Huang, Furong and Aloimonos, Yiannis},
  journal={arXiv preprint},
  year={2025}
}

@inproceedings{aloha,
  title={Learning Fine-Grained Bimanual Manipulation with Low-Cost Hardware},
  author={Zhao, Tony Z. and Kumar, Vikash and Levine, Sergey and Finn, Chelsea},
  booktitle={Robotics: Science and Systems (RSS)},
  year={2023}
}

@article{robomind2,
  title={RoboMIND 2.0: A Multimodal, Bimanual Mobile Manipulation Dataset for Generalizable Embodied Intelligence},
  author={Hou, Chengkai and Wu, Kun and Liu, Jiaming and Che, Zhengping and Wu, Di and Liao, Fei and Li, Guangrun and He, Jingyang and others},
  journal={arXiv preprint arXiv:2512.24653},
  year={2025}
}

@article{galaxea_g0,
  title={Galaxea Open-World Dataset and G0 Dual-System VLA Model},
  author={Jiang, Tao and Yuan, Tianyuan and Liu, Yicheng and Lu, Chenhao and Cui, Jianning and Liu, Xiao and Cheng, Shuiqi and Gao, Jiyang and others},
  journal={arXiv preprint arXiv:2509.00576},
  year={2025}
}

@article{molmoact2,
  title={MolmoAct2: Action Reasoning Models for Real-world Deployment},
  author={Fang, Haoquan and Duan, Jiafei and Clay, Donovan and Wang, Sam and Liu, Shuo and Huang, Weikai and Fan, Xiang and Tsai, Wei-Chuan and others},
  journal={arXiv preprint arXiv:2605.02881},
  year={2026}
}

@article{sekai,
  title={Sekai: A Video Dataset towards World Exploration},
  author={Li, Zhen and Li, Chuanhao and Mao, Xiaofeng and Lin, Shaoheng and Li, Ming and Zhao, Shitian and Xu, Zhaopan and Li, Xinyue and others},
  journal={arXiv preprint arXiv:2506.15675},
  year={2025}
}

@misc{egocentric100k,
  title={Egocentric-100K: 100,000 Hours of Real-World Egocentric Video from Factory Workers},
  author={{Build AI}},
  howpublished={\url{https://huggingface.co/datasets/builddotai/Egocentric-100K}},
  year={2026}
}

@article{kim24openvla,
    title={OpenVLA: An Open-Source Vision-Language-Action Model},
    author={{Moo Jin} Kim and Karl Pertsch and Siddharth Karamcheti and Ted Xiao and Ashwin Balakrishna and Suraj Nair and Rafael Rafailov and Ethan Foster and Grace Lam and Pannag Sanketi and Quan Vuong and Thomas Kollar and Benjamin Burchfiel and Russ Tedrake and Dorsa Sadigh and Sergey Levine and Percy Liang and Chelsea Finn},
    journal = {arXiv preprint arXiv:2406.09246},
    year={2024},
}

@article{liu2024rdt,
    title={RDT-1B: a Diffusion Foundation Model for Bimanual Manipulation},
    author={Liu, Songming and Wu, Lingxuan and Li, Bangguo and Tan, Hengkai and Chen, Huayu and Wang, Zhengyi and Xu, Ke and Su, Hang and Zhu, Jun},
    journal={arXiv preprint arXiv:2410.07864},
    year={2024}
}

@misc{ye2026worldactionmodelszeroshot,
  title         = {World Action Models are Zero-shot Policies},
  author        = {Ye, Seonghyeon and others},
  year          = {2026},
  eprint        = {2602.15922},
  archivePrefix = {arXiv},
  primaryClass  = {cs.RO},
  url           = {https://arxiv.org/abs/2602.15922},
}

@article{wan2025,
      title={Wan: Open and Advanced Large-Scale Video Generative Models}, 
      author={Team Wan and Ang Wang and Baole Ai and Bin Wen and Chaojie Mao and Chen-Wei Xie and Di Chen and Feiwu Yu and Haiming Zhao and Jianxiao Yang and Jianyuan Zeng and Jiayu Wang and Jingfeng Zhang and Jingren Zhou and Jinkai Wang and Jixuan Chen and Kai Zhu and Kang Zhao and Keyu Yan and Lianghua Huang and Mengyang Feng and Ningyi Zhang and Pandeng Li and Pingyu Wu and Ruihang Chu and Ruili Feng and Shiwei Zhang and Siyang Sun and Tao Fang and Tianxing Wang and Tianyi Gui and Tingyu Weng and Tong Shen and Wei Lin and Wei Wang and Wei Wang and Wenmeng Zhou and Wente Wang and Wenting Shen and Wenyuan Yu and Xianzhong Shi and Xiaoming Huang and Xin Xu and Yan Kou and Yangyu Lv and Yifei Li and Yijing Liu and Yiming Wang and Yingya Zhang and Yitong Huang and Yong Li and You Wu and Yu Liu and Yulin Pan and Yun Zheng and Yuntao Hong and Yupeng Shi and Yutong Feng and Zeyinzi Jiang and Zhen Han and Zhi-Fan Wu and Ziyu Liu},
      journal = {arXiv preprint arXiv:2503.20314},
      year={2025}
}

\let\clearpage\relax 
\end{document}